\documentclass{article}

   \usepackage[nonatbib,final]{neurips_2023}

\usepackage[utf8]{inputenc} 
\usepackage[T1]{fontenc}    
\usepackage{hyperref}       
\usepackage{url}            
\usepackage{booktabs}       
\usepackage{amsfonts}       
\usepackage{nicefrac}       
\usepackage{microtype}      
\usepackage{xcolor}         
\usepackage{graphicx}
\usepackage{subfigure}
\usepackage{parskip}
\usepackage{enumitem}
\usepackage{multirow}
\usepackage{wrapfig}
\usepackage{amsmath}
\usepackage{colortbl}
\usepackage{bbding}
\usepackage{caption2}
\usepackage[linesnumbered,ruled,vlined]{algorithm2e}
\usepackage{array}

\usepackage{booktabs}

\title{Sparse4D v3

Advancing End-to-End 3D Detection and Tracking}

\author{%
  Xuewu Lin, Zixiang Pei, Tianwei Lin, Lichao Huang, Zhizhong Su \\
  Horizon Robotics, Beijing, China\\
  \texttt{xuewu.lin@horizon.cc} \\
}

\begin{document}
\maketitle

\begin{abstract}
  In autonomous driving perception systems, 3D detection and tracking are the two fundamental tasks.
This paper delves deeper into this field, building upon the Sparse4D framework.
We introduce two auxiliary training tasks (Temporal Instance Denoising and Quality Estimation) and propose decoupled attention to make structural improvements, leading to significant enhancements in detection performance.
Additionally, we extend the detector into a tracker using a straightforward approach that assigns instance ID during inference, further highlighting the advantages of query-based algorithms. Extensive experiments conducted on the nuScenes benchmark validate the effectiveness of the proposed improvements. With ResNet50 as the backbone, we witnessed enhancements of 3.0\%, 2.2\%, and 7.6\% in mAP, NDS, and AMOTA, achieving 46.9\%, 56.1\%, and 49.0\%, respectively. Our best model achieved 71.9\% NDS and 67.7\% AMOTA on the nuScenes test set. Code will be released at \url{https://github.com/linxuewu/Sparse4D}.
\end{abstract}

\section{Introduction}

\begin{figure}[htbp]
\centering
\begin{minipage}[t]{0.49\textwidth}
\centering
\setcaptionwidth{6.6cm} 
\includegraphics[width=6.6cm]{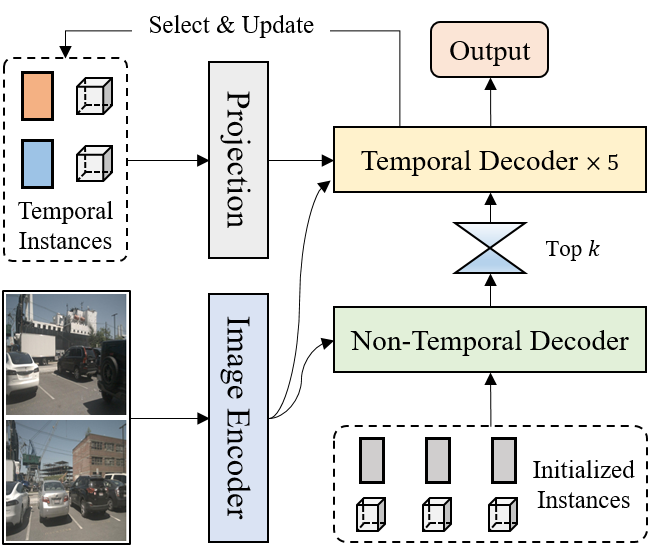}
\caption{Overview of Sparse4D framework, which input mutli-view video and output the perception results of all frames.}
\label{fig:overview}
\end{minipage}
\begin{minipage}[t]{0.49\textwidth}
\centering
\setcaptionwidth{6.2cm}
\includegraphics[width=6.2cm]{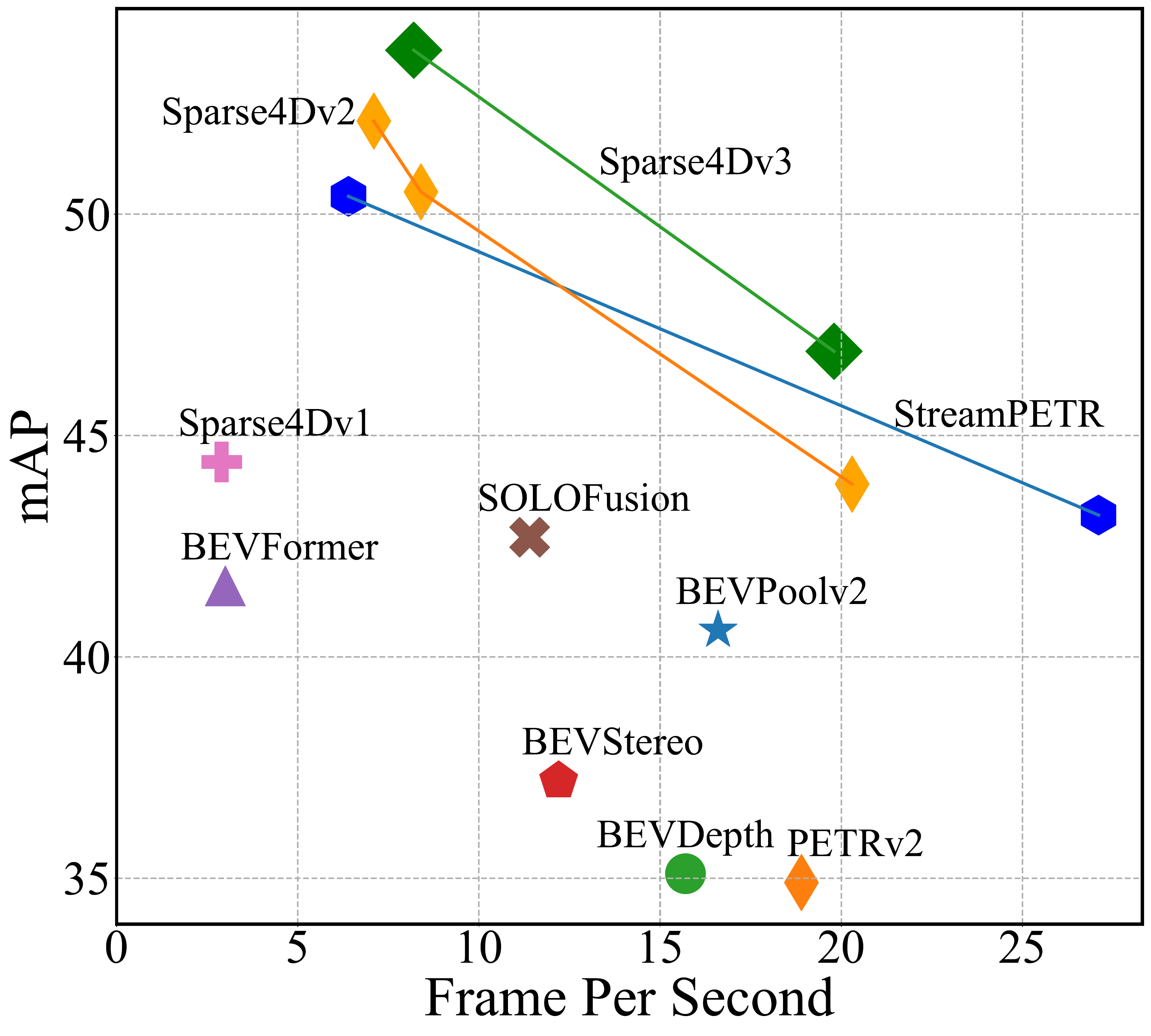}
\caption{Inference efficiency (FPS) - perception performance (mAP) on nuScenes validation dataset of different algorithms.}
\label{fig:fps-map}
\end{minipage}

\end{figure}

In the field of temporal multi-view perception research, sparse-based algorithms have seen significant advancements~\cite{wang2022detr3d,chen2022graphdetr3d,chen2022polardetr3d,mv2d,lin2022sparse4d,lin2023sparse4dv2}, reaching perception performance comparable to dense-BEV-based algorithms~\cite{bevformer,huang2021bevdet,huang2022bevdet4d,li2022bevdepth,li2022bevstereo,solofusion,bevformerv2,videobev} while offering several advantages:
1) Free view transform. These sparse methods eliminate the need for converting image space to 3D vector space. 
2) Constant computational load in detection head, which is irrelevant to perception distance and image resolution.
3) Easier implementation of integrating downstream tasks by end-to-end manner.
In this study, we select the sparse-based algorithm Sparse4Dv2~\cite{lin2022sparse4d,lin2023sparse4dv2} as our baseline for implementing improvements. The overall structure of the algorithm is illustrated in Figure~\ref{fig:overview}. The image encoder transforms multi-view images into multi-scale feature maps, while the decoder blocks leverage these image features to refine instances and generate perception outcomes.

To begin with, we observe that sparse-based algorithms encounter greater challenges in convergence compared to dense-based counterparts, ultimately impacting their final performance. This issue has been thoroughly investigated in the realm of 2D detection~\cite{li2022dn,zhang2022dino,codetr}, and is primarily attributed to the use of a one-to-one positive sample matching. This matching approach is unstable during the initial stages of training, and also results in a limited number of positive samples compared to one-to-many matching, thus reducing the efficiency of decoder training. Moreover, Sparse4D utilizes sparse feature sampling instead of global cross-attention, which further hampers encoder convergence due to the scarce positive samples. In Sparse4Dv2~\cite{lin2023sparse4dv2}, dense depth supervision has been introduced to partially mitigate these convergence issues faced by the image encoder. This paper primarily aims at enhancing model performance by focusing on the stability of decoder training. We incorporate the denoising task as auxiliary supervision and extend denoising techniques from 2D single-frame detection to 3D temporal detection. It not only ensures stable positive sample matching but also significantly increases the quantity of positive samples. Moreover, we introduce the task of quality estimation as auxiliary supervision.
This renders the output confidence scores more reasonable, refining the accuracy of detection result ranking and, resulting in higher evaluation metrics.
\begin{figure*}[]
    \centering
    \includegraphics[width=0.99\textwidth]{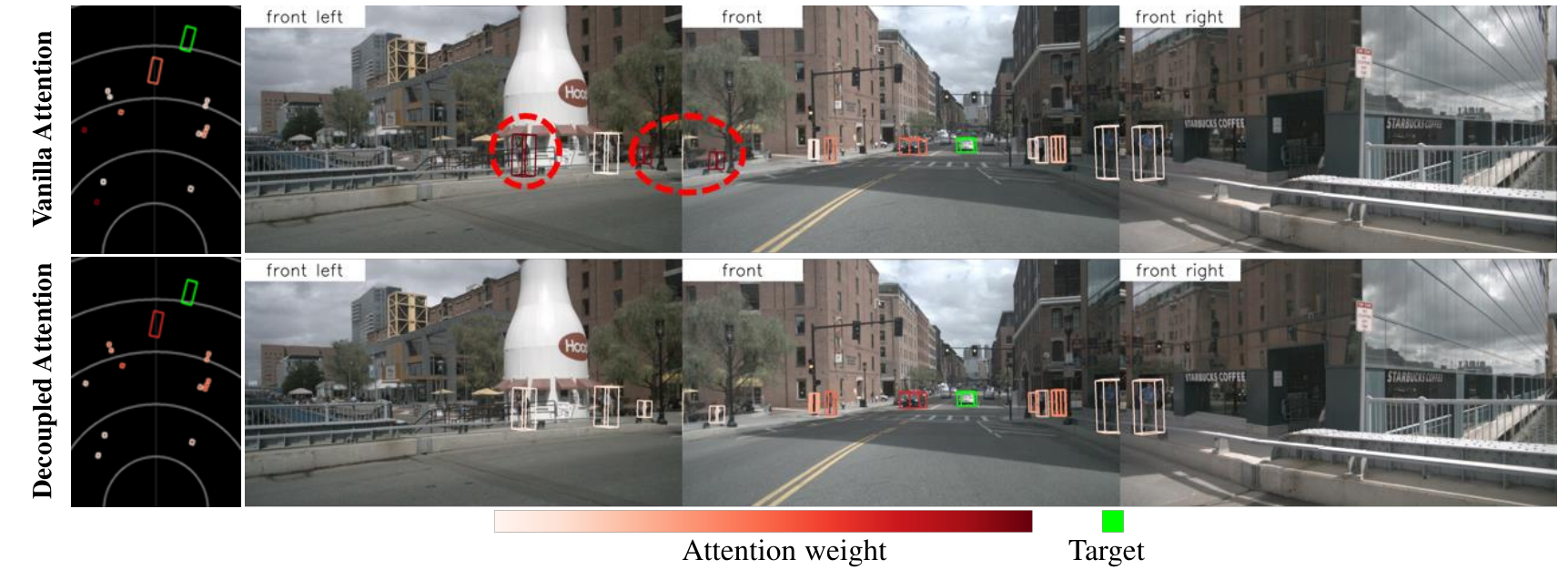}
    \caption{Visualizing Attention Weights in Instance Self-Attention: 1) The first row reveals attention weights in vanilla self-attention, where pedestrians in red circles show unintended correlations with the target vehicle (green box). 2) The second row displays attention weights in decoupled attention, effectively addressing the issue.}
    \label{fig:attention_weight}
\end{figure*}
Furthermore, we enhance the structure of the instance self-attention and temporal cross-attention modules in Sparse4D, introducing a decoupled attention mechanism designed to reduce feature interference during the calculation of attention weights. As depicted in Figure~\ref{fig:attention_weight},  when the anchor embedding and instance feature are added as the input for attention calculation, there are instances of outlier values in the resulting attention weights. This fails to accurately reflect the inter-correlation among target features, leading to an inability to aggregate the correct features. By replacing addition with concatenation, we significantly mitigate the occurrence of this incorrect phenomenon.
This enhancement shares similarities with Conditional DETR~\cite{meng2021conditional}. However, the crucial difference lies in our emphasis on attention among queries, as opposed to Conditional DETR, which concentrates on cross-attention between queries and image features. Additionally, our approach involves a distinct encoding methodology.
  
Finally, to advance the end-to-end capabilities of the perception system, we explore the integration of 3D multi-object tracking task into the Sparse4D framework, enabling the direct output of object motion trajectories. Unlike tracking-by-detection methods, we eliminate the need for data association and filtering, integrating all tracking functionalities into the detector.
Moreover, distinct from existing joint detection and tracking methods, our tracker requires no modification to the training process or loss functions. It does not necessitate providing ground truth IDs, yet achieves predefined instance-to-tracking regression.
Our tracking implementation maximally integrates the detector and tracker, requiring no modifications to the training process of the detector and no additional fine-tuning.
Our contributions can be summarized as follows:

(1) We propose Sparse4D-v3, a potent 3D perception framework with three effective  strategies: temporal instance denoising, quality estimation and decoupled attention.

(2) We extend Sparse4D into an end-to-end tracking model.

(3) We demonstrate the effectiveness of our improvements on nuScenes, achieving state-of-the-art performance in both detection and tracking tasks.

\section{Related Works}
\subsection{Improvements for End-to-End Detection}
DETR~\cite{DETR} lerverages the Transformer architecture~\cite{attention}, along with a one-to-one matching training approach, to eliminate the need for NMS and achieve end-to-end detection. DETR has led to a series of subsequent improvements. 
Deformable DETR~\cite{zhu2020deformabledetr} change global attention into local attention based on reference points, significantly narrowing the model's training search space and enhancing convergence speed. It also reduces the computational complexity of attention, facilitating the use of high-resolution inputs and multi-scale features within DETR's framework. Conditional-DETR~\cite{meng2021conditional} introduces conditional cross-attention, separating content and spatial information in the query and independently calculating attention weights through dot products, thereby accelerating model convergence. Building upon Conditional-DETR, Anchor-DETR\cite{wang2022anchorDETR} explicitly initializes reference points, serving as anchors. DAB-DETR~\cite{liu2021dab} further includes bounding box dimensions into the initialization of anchors and the encoding of spatial queries.
Moreover, many methods aim to enhance the convergence stability and detection performance of DETR from the perspective of training matching. DN-DETR~\cite{li2022dn} encodes ground truth with added noise as query input to the decoder, employing a denoising task for auxiliary supervision. Building upon DN-DETR, DINO~\cite{zhang2022dino} introduces noisy negative samples and proposes the use of Mixed Query Selection for query initialization, further improving the performance of the DETR framework. 
Group-DETR~\cite{chen2022groupdetr} replicates queries into multiple groups during training, providing more training samples. 
Co-DETR~\cite{codetr} incorporates dense heads during training, serving a dual purpose. It enables more comprehensive training of the backbone and enhances the training of the decoder by using the dense head output as a query.

DETR3D~\cite{wang2022detr3d} applies deformable attention to multi-view 3D detection, achieving end-to-end 3D detection with spatial feature fusion. The PETR series~\cite{liu2022petr,liu2022petrv2,streampetr}  introduce 3D position encoding, leveraging global attention for direct multi-view feature fusion and conducting temporal optimization. The Sparse4D series~\cite{lin2022sparse4d,lin2023sparse4dv2} enhance DETR3D in aspects like instance feature decoupling, multi-point feature sampling, temporal fusion, resulting in enhanced perceptual performance.

\subsection{Multi-Object Track}
Most multi-object tracking (MOT) methods use the tracking-by-detection framework. They rely on detector outputs to perform post-processing tasks like data association and trajectory filtering, leading to a complex pipeline with numerous hyperparameters that need tuning. These approaches do not fully leverage the capabilities of neural networks.
To integrate the tracking functionality directly into the detector, GCNet~\cite{lin2021global}, TransTrack~\cite{sun2020transtrack} and TrackFormer~\cite{meinhardt2022trackformer} utilize the DETR framework. They implement inter-frame transfer of detected targets based on track queries, significantly reducing post-processing reliance. MOTR~\cite{zeng2022motr} advances tracking to a fully end-to-end process. MOTRv3~\cite{yu2023motrv3} addresses the limitations in detection query training of MOTR, resulting in a substantial improvement in tracking performance. MUTR3D~\cite{zhang2022mutr3d} applies this query-based tracking framework to the field of 3D multi-object tracking.
These end-to-end tracking methods share some common characteristics:
(1) During training, they constrain matching based on tracking objectives, ensuring consistent ID matching for tracking queries and matching only new targets for detection queries.
(2) They use a high threshold to transmit temporal features, passing only high-confidence queries to the next frame.
Our approach diverges from existing methods. We don't need to modify detector training or inference strategies, nor do we require ground truth for tracking IDs.

\section{Methodology}

\label{Methodology}

The network structure and inference pipeline is depicted in Figure~\ref{fig:overview}, mirroring that of Sparse4Dv2~\cite{lin2023sparse4dv2}.
In this section, we will first introduce two auxiliary tasks:  Temporal Instance Denoising (Sec~\ref{TID}) and Quality Estimation (Sec~\ref{QE}). Following this, we present a straightforward enhancement to the attention module, termed decoupled attention (Sec~\ref{Decoupled Attention}). Finally, we outline how to leverage Sparse4D to achieve 3D MOT (Sec~\ref{tracking}).

\begin{figure*}[]
    \centering
    \includegraphics[width=0.99\textwidth]{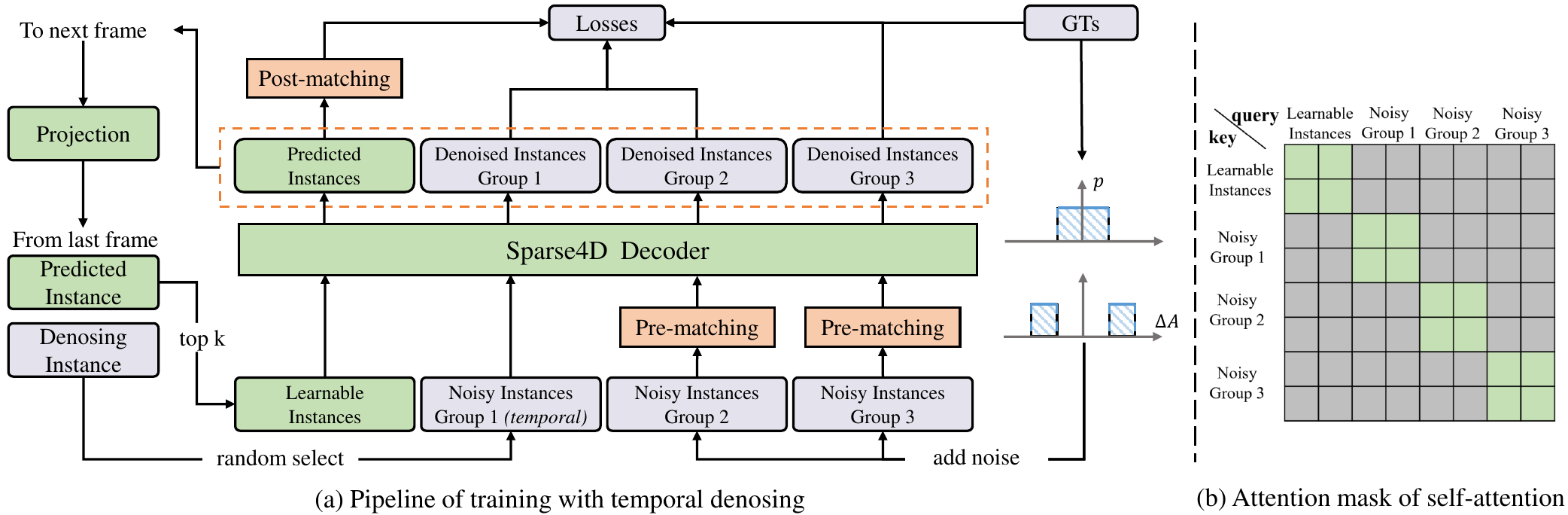}
    \caption{Illustration of Temporal Instance Denoising. (a) During the training phase, instances comprise two components: learnable and noisy. Noisy instances consist of both temporal and non-temporal elements. For noisy instances, we employ a pre-matching approach to allocate positive and negative samples—matching anchors with ground truth, while learnable instances are matched with predictions and ground truth . During the testing phase, only the green blocks in the diagram are retained. (b) Attention mask is employed to prevent feature propagation between groups, where gray indicates no attention between queries and keys, and green denotes the opposite.}
    \label{fig:temporal_denoising}
\end{figure*}
\subsection{Temporal Instance Denoising}
\label{TID}
In 2D detection, introducing a denoising task proves to be an effective approach for improving both model convergence stability and detection performance. In this paper, we extend the fundamental 2D single-frame denoising to 3D temporal denoising. Within Sparse4D, instances (referred to as queries) are decoupled into implicit instance features and explicit anchors. During the training process, we initialize two sets of anchors. One set comprises anchors uniformly distributed in the detection space, initialized using the k-means method, and these anchors serve as learnable parameters. The other set of anchors is generated by adding noise to ground truth (GT), as illustrated in Equation (\ref{eq:add_noise1},\ref{eq:add_noise2}), specifically tailored for 3D detection tasks.
\begin{gather}
A_{gt}=\left\{(x, y, z, w, l, h, yaw, v_{xyz})_i \;|\; i \in \mathbb{Z}_{N} \right\} \label{eq:add_noise1}\\
A_{noise} = \left\{A_{i} + \Delta A_{i,j,k} \; |\; i \in \mathbb{Z}_{N}, j \in \mathbb{Z}_{M}, k \in \mathbb{Z}_{2}\right\}
\label{eq:add_noise2}
\end{gather}

Here, $\mathbb{Z}_{X}$ represents the set of integers between 1 and $X$. $N$ denotes the number of GT, while $M$ represents the group number of noising instances. In this context, $\Delta A$ signifies random noise, where $\Delta A_{i,j,1}$ and $\Delta A_{i,j,2}$ follow uniform random distributions within the range $(-x, x)$ and $(-2x, -x) \cup (x, 2x)$, respectively.
In DINO-DETR~\cite{zhang2022dino}, which categorizes samples generated by $\Delta A_{i,j,1}$ as positive and those from $\Delta A_{i,j,2}$ as negative, there is a potential risk of mis-assignment, as $\Delta A_{i,j,2}$ may be closer to the ground truth. To entirely mitigate any ambiguity, we employ bipartite graph matching for each group of $A_{noise}$ and $A_{gt}$ to determine positive and negative samples.

Furthermore, we extend the aforementioned single-frame noisy instances through temporal propagation to better align with the sparse recurrent training process. During each frame's training, we randomly select $M'$ groups from the noisy instances to project onto the next frame. The temporal propagation strategy aligns with that of non-noisy instances—anchors undergo ego pose and velocity compensation, while instance features serve as direct initializations for the features of the subsequent frame.

It's important to note that we maintain the mutual independence of each group of instances, and no feature interaction occurs between noisy instances and normal instances. This is different from DN-DETR~\cite{li2022dn}, as shown in Figure~\ref{fig:temporal_denoising}(b). This approach ensures that within each group, a ground truth is matched to at most one positive sample, effectively avoiding any potential ambiguity.

\subsection{Quality Estimation}
\label{QE}

Existing sparse-based methods primarily estimate classification confidence for positive and negative samples to measure alignment with ground truth. The optimization goal is to maximize the classification confidence of all positive samples. However, there is significant variation in matching quality among different positive samples. Consequently, classification confidence is not an ideal metric for evaluating the quality of predicted bounding boxes.
To facilitate the network in understanding the quality of positive samples, accelerating convergence on one hand and rationalizing the prediction ranking on the other, we introduce the task of prediction quality estimation.
For the 3D detection task, we define two quality metrics: centerness and yawness, with the following formulas.
\begin{gather}
    C = \exp (-\lVert [x, y, z]_{pred} - [x, y, z]_{gt} \rVert _{2}) \label{eq:centerness} \\
    Y = [\sin yaw, \cos yaw]_{pred} \cdot [\sin yaw, \cos yaw]_{gt} \label{eq:yawness}
\end{gather}

While network outputs classification confidence, it also estimates centerness and yawness. Their respective loss functions are defined as cross-entropy loss and focal loss~\cite{lin2017focal}, as depicted in the following equation.

\begin{equation}
    L = \lambda_{1} \textbf{CE}(Y_{pred}, Y) + \lambda_{2} \textbf{Focal}(C_{pred}, C)
\end{equation}

\subsection{Decoupled Attention}
\label{Decoupled Attention}
\begin{figure*}[ht]
    \centering
    \includegraphics[width=0.99\textwidth]{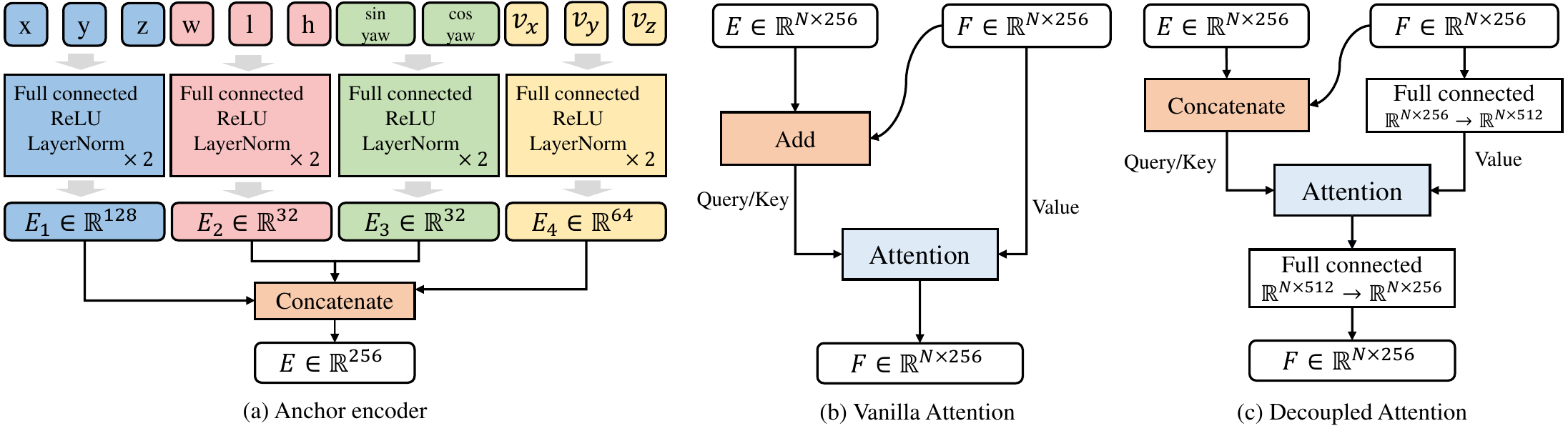}
    \caption{Architecture of the anchor encoder and attention. We independently conduct high-dimensional feature encoding on multiple components of the anchor and subsequently concatenate them. This approach leads to lower computational and parameter overhead compared to the original Sparse4D. $E$ and $F$ represent anchor embedding and instance feature, respectively.}
    \label{fig:decouple_attn}
\end{figure*}

As mentioned in the introduction, we make simple improvements to the anchor encoder, self-attention, and temporal cross-attention in Sparse4Dv2. The architecture is illustrated in Figure~\ref{fig:decouple_attn}. The design principle is to combine features from different modalities in a concatenated manner, as opposed to using an additive approach.
There are some differences compared to Conditional DETR~\cite{meng2021conditional}. Firstly, we make improvements in the attention between of queries instead of the cross-attention between query and image features; the cross-attention still utilizes deformable aggregation from Sparse4D. Additionally, instead of concatenating position embedding and query feature at the single-head attention level, we make modifications externally at the multi-head attention level, providing the neural network with greater flexibility.

\subsection{Extend to Tracking}
\label{tracking}
In the framework of Sparse4Dv2, the temporal modeling adopts a recurrent form, projecting instances from the previous frame onto the current frame as input. The temporal instances are similar to the tracking queries in query-based trackers, with the distinction that the tracking queries are constrained by a higher threshold, representing highly confident detection results. In contrast, our temporal instances are numerous, and most of them may not accurately represent detected objects in previous frames.

To extend from detection to multi-object tracking within the Sparse4Dv2 framework, we directly redefine the instance from a detection bounding box to a trajectory. A trajectory includes an ID and bounding boxes for each frame.
Due to the setting of a large number of redundant instances, many instances may not be associated with a precise target and do not be assigned a definite ID.
Nevertheless, they can still be propagated to the next frame. Once the detection confidence of an instance surpasses the threshold $T$, it is considered to be locked onto a target and assigned an ID, which remains unchanged throughout temporal propagation.
Therefore, achieving multi-object tracking is as simple as applying an ID assignment process to the output perception results. The lifecycle management during tracking is seamlessly handled by the top-k strategy in Sparse4Dv2, requiring no additional modifications. Specifics can be referred to in Algorithm~\ref{alg:tracking}. In our experiments, we observe that the trained temporal model demonstrates excellent tracking characteristics without the need for fine-tuning with tracking constraints.

\IncMargin{1em}
\begin{algorithm}[H]
\caption{Tracking Pipeline of Sparse4Dv3}
\label{alg:tracking}

\KwIn{

\textbf{1) Sensor Data} $D$,
\textbf{2) Temporal Instances} $I_{t}=\left\{(c, a, \mathrm{id})_{i} | i \in \mathbb{Z}_{N_{t}} \right\}$,
\textbf{3) Current Instances} $I_{cur}=\left\{a_{i} | i \in \mathbb{Z}_{N_{cur}} \right\}$.
The terms $c$, $a$ and $\mathrm{id}$ correspond to confidence, anchor box, and object ID, respectively, where the $\mathrm{id}$ may be empty.
}

\KwOut{

\textbf{1) Perception Results} $R=\left\{(c, a, \mathrm{id})_{i} | i \in \mathbb{Z}_{N_{r}}\right\}$,
\textbf{2) Updated} $I'_{t}=\left\{(c, a, \mathrm{id})_{i} | i \in \mathbb{Z}_{N_{t}} \right\}$.
}
Define the confidence threshold $T$.

Define the confidence decay scale $S$.

Initialize $R=\left\{\right\}$.

Forward the model $R_{det}=\textbf{Model}(D, I_{t}, I_{cur})=\left\{(c', a')_{i} | i \in \mathbb{Z}_{N_{t}+N_{cur}}\right\}$.

\For{$i$ {\rm \textbf{from} 1} {\rm \textbf{to}} $N_{t}+N_{cur}$}{
    \If{$c'_{i} \ge T$}{
        \If{
            $i > N_{t}$ {\rm\textbf{or}} $\mathrm{id}_{i}$ is empty
        }{Generate the new $\mathrm{id}_i$.}
          Put $(c'_i, a'_i, \mathrm{id}_{i})$ into $R$.
    }
    \If{$i \le N_{t}$}
    {
        Update $c'_i = \mathrm{max}(c'_i, c_i \times S)$.
    }
}
Select $N_{t}$ instances with the highest confidence $c'$ from the $R_{det}$ to serve as $I'_{t}$.

\textbf{Return} $R$ and $I'_{t}$.
\end{algorithm}
\DecMargin{1em}

\section{Experiment}
\label{Experiment}
\subsection{Benckmark}
To validate the effectiveness of Sparse4Dv3, we employ the nuScenes benchmark, a dataset containing 1000 scenes. The distribution for training, validation, and testing is 700, 150, and 150 scenes, respectively. Each scene features a 20-second video clip at 2 frames per second (FPS) and includes 6 viewpoint images. Alongside 3D bounding box labels, the dataset provides data on vehicle motion states and camera parameters.
For detection performance evaluation, a comprehensive approach considers metrics such as mean Average Precision (mAP), mean Average Error of Translation (mATE), Scale (mASE), Orientation (mAOE), Velocity (mAVE), Attribute (mAAE), and nuScenes Detection Score (NDS), where NDS represents a weighted average of the other metrics.
For tracking model evaluation, key metrics include Average Multi-Object Tracking Accuracy (AMOTA), Average Multi-Object Tracking Precision (AMOTP), Recall, and ID Switch (IDS). For a detailed understanding, please refer to \cite{caesar2020nuscenes,leal2015motchallenge}.
  
\subsection{Implementation Details}
Following Sparse4Dv2~\cite{lin2023sparse4dv2}, unless explicitly stated otherwise, our head utilizes a 6-layer decoder, comprising 900 instances and $N_{t}=600$ temporal instances, with an embedding dimension of 256.
Additionally, it incorporates 7 fixed keypoints and 6 learnable keypoints. In Algorithm~\ref{alg:tracking}, the parameters $T$ and $S$ are set to 0.25 and 0.6, respectively. The number of groups for denoising instances $M$ is 5, and among them, 3 groups are randomly selected as temporal denoising instances.

We train the model for 100 epochs using the AdamW optimizer without the CBGS~\cite{zhu2019cbgs}, and there is no need for any fine-tuning on the tracking task.
Like most methods~\cite{solofusion,lin2023sparse4dv2,streampetr}, we employ sequential iteration approach for training. Each training step takes input data from a single frame and instances cached from historical frames. The training duration and GPU memory consumption of the temporal model are similar to those of the single-frame model, allowing us to efficiently train the temporal model. In addition to the temporal instance denoising and quality estimation tasks introduced in this paper, we also incorporate dense depth regression~\cite{lin2023sparse4dv2} as an auxiliary task to enhance the stability of model training.

\subsection{Main Results}
To better control variables, we conduct thorough comparative experiments on the validation dataset, as shown in Table~\ref{tab:detectionval}. In the first experimental setup, we use ResNet50~\cite{resnet} as the backbone, initialized with parameters from supervised training on ImageNet-1k~\cite{krizhevsky2017imagenet}. The image size is set to $256\times 704$. These parameters have relatively lower requirements on GPU memory and training time, facilitating experimental iteration. In this configuration, Sparse4Dv3 improve mAP and NDS by 3.0\% and 2.2\%, respectively.
In the second experimental setup, ResNet101 is employed as the backbone, and the image size is doubled to $512\times 1408$ to evaluate the model's performance on larger images. Sparse4Dv3 achieves state-of-the-art performance in this configuration as well, with mAP and NDS improving by 3.2\% and 2.9\%, respectively. Moreover, compared to Sparse4Dv2, the inference speed remains almost unchanged. In the $512\times 1408$ configuration, our inference speed still surpasses that of StreamPETR, which utilizes global attention.

\noindent\textbf{3D detection on test set.}
Furthermore, we evaluate the model's performance on the nuScenes test dataset, as shown in Table~\ref{tab:detectiontest}. To maintain consistent configurations with most algorithms, we employ VoVNet-99~\cite{vovnet} as the backbone, with pretrained weights from DD3D~\cite{DD3D}, and set the image size to $640\times 1600$. On the test dataset, Sparse4Dv3 achieves optimal performance in both metrics, with mAP and NDS improving by 1.3\% and 1.8\%, respectively. Importantly, the distance error (mATE) performance of our sparse-based algorithm significantly surpasses that of the dense-BEV-based algorithms. This is primarily attributed to the stability in confidence ordering achieved through our adopted quality estimation, leading to a substantial improvement in mATE.

\noindent\textbf{Multi-object tracking 3D.}
We directly evaluate the models using MOT3D metrics in both Table~\ref{tab:detectionval} and Table~\ref{tab:detectiontest} without extra fine-tuning. As shown in Table~\ref{tab:trackingval}, on the validation evaluation set, Sparse4Dv3 significantly outperforms existing methods across all tracking metrics, whether they are end-to-end or non-end-
to-end methods. Compared to the SOTA solution DORT~\cite{lian2023dort}, under the same configuration, our AMOTA is 6.6\% higher (0.490 vs 0.424). In comparison to the end-to-end solution DQTrack, our AMOTA improves by 16.0\% (0.567 vs 0.407), and the ID switch is reduced by 44.5\% (557 vs 1003). Table~\ref{tab:trackingtest} presents the evaluation results on the test dataset, where Sparse4Dv3 achieves state-of-the-art performance across metrics such as IDS, recall, MOTAR, MOTA, and MOTP.

\begin{table*}
  \centering
  \scriptsize
  \setlength{\tabcolsep}{1.5mm}{
  \begin{tabular}{@{}l|c|c|cccccc|c|c@{}}
    \toprule
    Method & Backbone & Image size &mAP$\uparrow$ & mATE$\downarrow$ & mASE$\downarrow$ & mAOE$\downarrow$ & mAVE$\downarrow$ & mAAE$\downarrow$ & NDS$\uparrow$ & FPS$\uparrow$ \\
    \midrule
    BEVPoolv2~\cite{huang2022bevpoolv2} & ResNet50 & $256\times 704$ & 0.406 & 0.572 & 0.275 & 0.463 & 0.275 & 0.188 & 0.526 & 16.6 \\
    BEVFormerv2~\cite{bevformerv2} & ResNet50 &- & 0.423 & 0.618 & 0.273 & \textbf{0.413} & 0.333 & 0.188 & 0.529 & - \\
    SOLOFusion~\cite{solofusion} & ResNet50 & $256\times 704$ & 0.427 & 0.567 & 0.274 & 0.511 & 0.252 & 0.181 & 0.534 & 11.4 \\
    VideoBEV~\cite{videobev} & ResNet50 & $256\times 704$ & 0.422 & 0.564 & 0.276 & 0.440 & 0.286 & 0.198 & 0.535 & -\\
    StreamPETR~\cite{streampetr} & ResNet50 & $256\times 704$ & 0.432 & 0.609 & \textbf{0.270} & 0.445 & 0.279 & 0.189 & 0.537 & \textbf{26.7} \\
    Sparse4Dv2 & ResNet50 & $256\times 704$ & 0.439 & 0.598 & \textbf{0.270} & 0.475 & 0.282 & \textbf{0.179} & 0.539 & 20.3 \\
    \specialrule{0em}{1pt}{1pt}
    \rowcolor{gray!30} Sparse4Dv3 & ResNet50 & $256\times 704$ & \textbf{0.469} & \textbf{0.553} & 0.274 & 0.476 & \textbf{0.227} & 0.200 & \textbf{0.561} & 19.8\\
    \specialrule{0em}{2pt}{2pt}
    \bottomrule
    \specialrule{0em}{2pt}{2pt}
    
    BEVDepth~\cite{li2022bevdepth} & ResNet101 & $512\times 1408$ & 0.412 & 0.565 & 0.266 & 0.358 & 0.331 & 0.190 & 0.535 & - \\
    Sparse4D~\cite{lin2022sparse4d} & Res101-DCN & $640\times 1600$ & 0.444 & 0.603 & 0.276 & 0.360 & 0.309 & \textbf{0.178} & 0.550 & 2.9 \\
    SOLOFusion & ResNet101 & $512\times 1408$ & 0.483 & \textbf{0.503} & 0.264 & 0.381 & 0.246 &  0.207 & 0.582 & - \\
    StreamPETR$^{\dagger}$ & ResNet101 & $512\times 1408$ & 0.504 & 0.569 & 0.262 & 0.315 & 0.257 & 0.199 & 0.592 & 6.4 \\
    Sparse4Dv2$^{\dagger}$ & ResNet101 & $512\times 1408$ & 0.505 & 0.548 & 0.268 & 0.348 & 0.239 & 0.184 & 0.594 & \textbf{8.4} \\
    \specialrule{0em}{1pt}{1pt}
    \rowcolor{gray!30} Sparse4Dv3$^{\dagger}$ & ResNet101 & $512\times 1408$ & \textbf{0.537}&0.511&\textbf{0.255}&\textbf{0.306}&\textbf{0.194}&0.192&\textbf{0.623}&8.2\\
    \bottomrule
  \end{tabular}}
  \caption{Results of 3D detection on nuScenes validation dataset. $\dagger$ indicates to use pre-trained weights from the nuImage dataset.}
  \label{tab:detectionval}
\end{table*}

\begin{table*}
  \centering
  \scriptsize
  \setlength{\tabcolsep}{1.5mm}{
  \begin{tabular}{@{}l|c|cccccc|c@{}}
    \toprule
    Method & Backbone & mAP$\uparrow$ & mATE$\downarrow$ & mASE$\downarrow$ & mAOE$\downarrow$ & mAVE$\downarrow$ & mAAE$\downarrow$ & NDS$\uparrow$ \\
    \midrule
    Sparse4D & VoVNet-99 & 0.511 & 0.533 & 0.263 & 0.369 & 0.317 & 0.124 & 0.595\\
    HoP-BEVFormer~\cite{hop} & VoVNet-99 & 0.517 & 0.501 & 0.245 & 0.346 & 0.362 & \textbf{0.105} & 0.603 \\ 
    SOLOFusion & ConvNeXt-B~\cite{liu2022convnet} & 0.540 & 0.453 & 0.257 & 0.376 & 0.276 & 0.148 & 0.619 \\
    BEVFormerv2 & InternImage-B~\cite{wang2022internimage} & 0.540 & 0.488 & 0.251 & 0.335 & 0.302 & 0.122 & 0.620 \\
    VideoBEV & ConvNeXt-B & 0.554 & 0.457 & 0.249 & 0.381 & 0.266 & 0.132 & 0.629 \\
    StreamPETR & VoVNet-99 & 0.550 & 0.479 & 0.239 & 0.317 & 0.241 & 0.119 & 0.636 \\
    Sparse4Dv2 & VoVNet-99 & 0.557 & 0.462 & 0.238 & 0.328 & 0.264 & 0.115 & 0.638\\
    \rowcolor{gray!30} Sparse4Dv3 & VoVNet-99 & \textbf{0.570} & \textbf{0.412} & \textbf{0.236} & \textbf{0.312} & \textbf{0.210} & 0.117 & \textbf{0.656}\\
    \bottomrule
  \end{tabular}}
  \caption{Results of 3D detection on nuScenes test dataset. All VoVNet-99 in the table are initialized with weights from DD3D~\cite{DD3D}.}
  \label{tab:detectiontest}
\end{table*}

\begin{table*}
  \centering
  \scriptsize
  \setlength{\tabcolsep}{1.5mm}{
  \begin{tabular}{@{}l|c|c|c|ccccccc@{}}
    \toprule
    Method & E2E & Backbone & Image size & AMOTA$\uparrow$ & AMOTP$\downarrow$ & IDS$\downarrow$ & Recall$\uparrow$ & MOTA$\uparrow$ & MOTP$\downarrow$   \\
    \midrule
    QTrack~\cite{qtrack} &  & ResNet50 & $256\times 704$ & 0.347 & 1.347 & 944 & 0.426 &0.426 & 0.722 \\
    DORT~\cite{lian2023dort} &  & ResNet50 & $256\times 704$ & 0.424 & 1.264 & - & 0.492& -& -\\
    \rowcolor{gray!30} Sparse4Dv3 &\checkmark  & ResNet50 & $256\times 704$ & \textbf{0.490} & \textbf{1.164} & \textbf{430} & \textbf{0.574}  & \textbf{0.436} & \textbf{0.660}\\ 
    \bottomrule
    DQTrack~\cite{dqtrack}  & \checkmark & ResNet101 & $512\times 1408$ & 0.407 & 1.318 & 1003 & -& -& -\\
    SRCN3D~\cite{shi2022srcn3d} & &  VoVNet-99 & $900\times 1600$& 0.439 & 1.280 & - & 0.545&  -& -\\
    QTrack &  & VoVNet-99 & $640\times 1600$ & 0.511 & 1.090 & 1144 & 0.585& 0.465 & - \\
    \rowcolor{gray!30} Sparse4Dv3 &\checkmark  & ResNet101 & $512\times 1408$ & \textbf{0.567} & \textbf{1.027} & \textbf{557} & \textbf{0.658} & \textbf{0.515} & \textbf{0.621}\\ 
    \bottomrule

  \end{tabular}}
  \caption{Results of 3D multi-object tracking on nuScenes val dataset. "E2E" stands for whether the model is an end-to-end detection and tracking model. The model of Sparse4Dv3 in this table is same as in Table~\ref{tab:detectionval}.}
  \label{tab:trackingval}
\end{table*}

\begin{table*}
  \centering
  \scriptsize
  \setlength{\tabcolsep}{1.5mm}{
  \begin{tabular}{@{}l|c|c|ccccccc@{}}
    \toprule
    Method & E2E & Backbone & AMOTA$\uparrow$ & AMOTP$\downarrow$ & IDS$\downarrow$ & Recall$\uparrow$ & MOTAR$\uparrow$ & MOTA$\uparrow$ & MOTP$\downarrow$   \\
    \midrule
    MUTR3D~\cite{zhang2022mutr3d} & \checkmark & ResNet101 & 0.270 & 1.494 & 6018 & 0.411 &0.643 & 0.245 & 0.709 \\
    SRCN3D~\cite{shi2022srcn3d} &  &  VoVNet-99 & 0.398 & 1.317 & 4090 & 0.538 & 0.702 & 0.359 & 0.709 \\
    DQTrack~\cite{dqtrack}& \checkmark&  VoVNet-99 & 0.523 & 1.096 & 1204 & 0.622 & -  & 0.444 & 0.649\\
    QTrack-StP~\cite{qtrack,streampetr} &  &  ConvNext-B & 0.566 & 0.975 & 784 &  0.650 & 0.711 & 0.460 & 0.576 \\
    DORT~\cite{lian2023dort} &  & ConvNext-B & \textbf{0.576} & \textbf{0.951} & 774 & 0.634 & 0.771 & 0.484 & 0.536 \\
    \rowcolor{gray!30} Sparse4Dv3 &\checkmark & VoVNet-99 & 0.574 & 0.970 & \textbf{669} & \textbf{0.669} &  \textbf{0.779} &\textbf{0.521} &\textbf{0.525}\\
    \bottomrule

  \end{tabular}}
  \caption{Results of 3D multi-object tracking on nuScenes test dataset. "E2E" stands for whether the model is an end-to-end detection and tracking model. "QTrack-StP" denotes QTrack with StreamPETR as detector. The model of Sparse4Dv3 in this table is same as in Table~\ref{tab:detectiontest}.}
  \label{tab:trackingtest}
\end{table*}

\subsection{Ablation Studies}
\begin{figure*}[ht]
    \centering
    \includegraphics[width=0.99\textwidth]{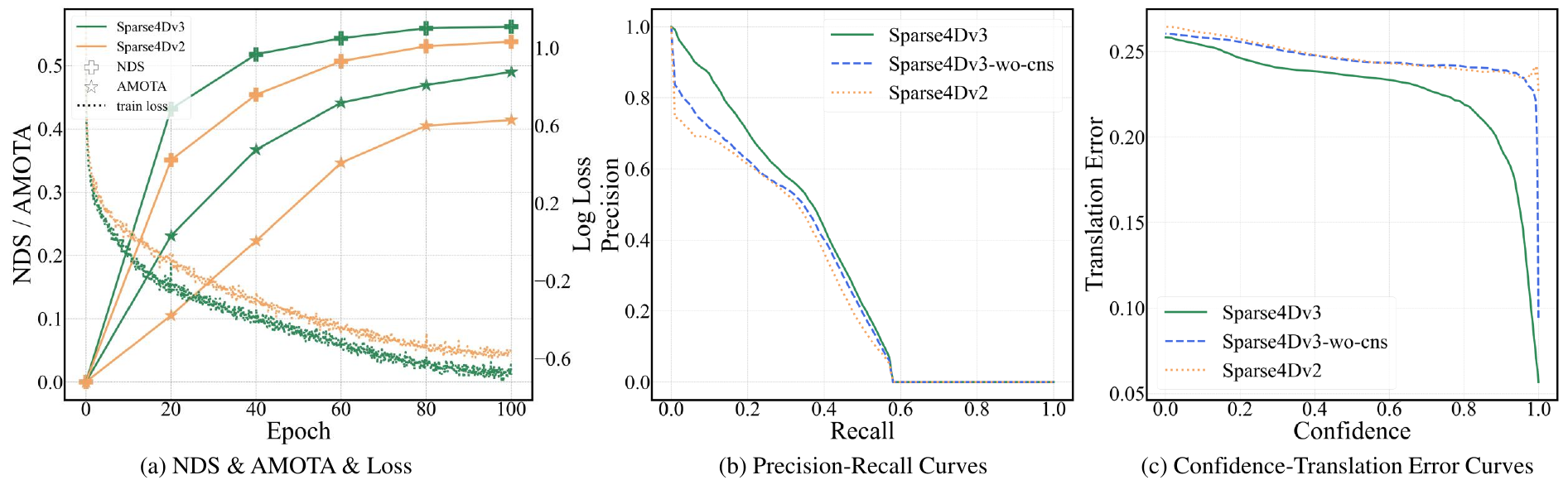}
    \caption{(a) Comparison of evaluation metrics and loss during the training process of sparse4dv2 and v3. (b, c) Comparison of Precision-Recall and Confidence-Translation error for the pedestrian category before and after the incorporation of centerness, using a threshold of translation error=0.5m for precision and recall.}
    \label{fig:curves}
\end{figure*}

We conduct ablation experiments to assess the improvements in this paper. Under controlled variables, we incrementally introduce individual enhancements, and the results are shown in Table~\ref{tab:ablation}. Specifically, denoising demonstrates significant improvements across various metrics. Single-frame denoising and temporal denoising contribute to increases of 0.8\% and 0.4\% in mAP, as well as 0.9\% and 0.6\% in NDS, respectively. Decoupled attention primarily enhances the model's mAP and mAVE, with improvements of 1.1\% and 1.9\%, respectively. Centerness aligns with its intended design, notably reducing distance error by 1.8\%. However, it has a negative impact on orientation estimation error, which is partially mitigated by the introduction of yawness. The combination of the two quality estimations results in improvements of 0.8\% in mAP, 2.8\% in mATE, and 1.9\% in mAVE. Figure~\ref{fig:curves} (a) illustrates the training curves of loss and evaluation metrics for Sparse4Dv2 and Sparse4Dv3. It can be observed that Sparse4Dv3 shows a significant improvement in both final convergence and convergence speed compared to Sparse4Dv2.

We further validate the fundamental reason for the impact of centerness on model performance. We believe that the confidence trained through the classification loss in one-to-one matching does not effectively reflect the quality of the detection result. A high-confidence box does not necessarily correspond to a closer ground truth. After incorporating centerness, we use centerness multiplied by confidence as the score for each detection result, making the ranking of detection results more accurate. As shown in Figure~\ref{fig:curves} (b, c), when centerness is not used, the model's accuracy is still not very high in situations of low recall and high thresholds (e.g., recall $\le$ 0.1 or threshold $\ge$ 0.95), and the translation error is relatively large. The introduction of centerness significantly mitigates this phenomenon.

\begin{table*}
  \centering
  \scriptsize
  \setlength{\tabcolsep}{1.5mm}{
  \begin{tabular}{@{}lccccccc|cc@{}}
    \toprule
    Setting &mAP$\uparrow$ & mATE$\downarrow$ & mASE$\downarrow$ & mAOE$\downarrow$ & mAVE$\downarrow$ & mAAE$\downarrow$ & NDS$\uparrow$ & AMOTA$\uparrow$  & AMOTR$\downarrow$\\
    \midrule
   Sparse4Dv2 & 0.439 & 0.598 & 0.270 & 0.475 & 0.282 & 0.179 & 0.539 & 0.414 & 1.268\\
   \midrule
   $+$ single-frame denoising & 0.447 & 0.586 & 0.267 & 0.455 &  0.257 & 0.187 & 0.548 & 0.445 & 1.230\\
   $+$ decoupled attention & 0.458 & 0.599 & 0.268 & 0.481 & 0.238 & 0.192 & 0.551 & 0.472 & 1.195 \\
   $+$ temporal denoising & 0.462 & 0.581 &  0.269 & 0.454 & 0.246 & 0.189 & 0.557 & 0.457 & 1.191 \\
   $+$ centerness & 0.463 &0.563 &0.265 & 0.517 &0.221 &0.208 &0.554 & 0.466 & 1.180\\
   $+$ yawness & 0.469 & 0.553 & 0.274 & 0.476 & 0.227 & 0.200 & 0.561 & 0.490 &  1.164 \\
   \midrule
   \rowcolor{gray!30} $=$ Sparse4Dv3 & 
   \textbf{\textcolor[RGB]{0,150,80}{$+$0.030}} & 
   \textbf{\textcolor[RGB]{0,150,80}{$-$0.045}}& 
   \textbf{{\color{red} $+$0.004}} & 
   \textbf{{\color{red} $+$0.001}} &
   \textbf{\textcolor[RGB]{0,150,80}{$-$0.055}}&
   \textbf{{\color{red} $+$0.021}} &
   \textbf{\textcolor[RGB]{0,150,80}{$+$0.022}}&
   \textbf{\textcolor[RGB]{0,150,80}{$+$0.076}}&
   \textbf{\textcolor[RGB]{0,150,80}{$-$0.104}}\\
   \bottomrule
  \end{tabular}}
  \caption{Ablation Experiments. In the last row, \textbf{\textcolor[RGB]{0,150,80}{green}} font indicates an improvement in the metric, while \textbf{{\color{red}red}} font indicates the opposite.}
  \label{tab:ablation}
\end{table*}

\subsection{Cloud-Based Performance Boost}
With ample computational power available in cloud-based systems, it is common to leverage larger computational resources to achieve optimal performance. Therefore, we employ two measures to further unleash the potential of Sparse4D, including feature fusion with future frames and a larger, better-pretrained backbone.
Firstly, we adopt the multi-frame sampling approach from Sparse4Dv1~\cite{lin2022sparse4d} to fuse features from future frames. With the inclusion of features from the next 8 frames (2 FPS), there is a significant improvement in model performance, particularly with a 5.67\% reduction in mAVE and a 3.23\% increase in NDS. Additionally, following the approach in StreamPETR~\cite{streampetr}, we experiment with using EVA02~\cite{fang2023eva02} as the backbone. EVA02 has undergone thorough pretraining, and its feature extraction is rich in semantic information, offering stronger generalization and aiding in model classification. Compared to ResNet101, EVA02-Large leads to a 5.98\% increase in mAP. By incorporating both EVA02 and future frames, we achieve a remarkable mAP of 0.682, NDS of 0.719, and AMOTA of 0.677 on the nuScenes test dataset. this achievement even surpasses certain metrics (NDS and mAVE) compared to detection models that utilize lidar, such as TransFusion~\cite{bai2022transfusion}.

\begin{table*}
  \centering
  \scriptsize
  \setlength{\tabcolsep}{1.5mm}{
  \begin{tabular}{@{}l|cc|ccccc|cc@{}}
    \toprule
    Method &Backbone & Future&mAP$\uparrow$ & mATE$\downarrow$  & mAOE$\downarrow$ & mAVE$\downarrow$ & NDS$\uparrow$ & AMOTA$\uparrow$  & AMOTR$\downarrow$\\
    \midrule
    TransFusion-L~\cite{bai2022transfusion} & DLA34\&VoxelNet & 0 & 0.655 &0.256 &0.351 &0.278 & 0.702 & 0.686 & 0.529\\
    TransFusion-CL~\cite{bai2022transfusion} & DLA34\&VoxelNet & 0 & \textbf{0.689} & \textbf{0.259} & 0.359 &0.288 & 0.717 & \textbf{0.718} & \textbf{0.551}\\
    \midrule
    \rowcolor{gray!30} Sparse4Dv3 & ResNet101 & 0 & 0.570 & 0.411 & 0.303 & 0.202 & 0.658 & 0.571 & 0.995\\
    \rowcolor{gray!30} Sparse4Dv3 & ResNet101 & 8 & 0.613 & 0.371 & 0.285 & 0.144 & 0.690 & 0.608 & 0.876\\
    \rowcolor{gray!30} Sparse4Dv3 & EVA02-Large~\cite{fang2023eva02} & 0 & 0.630 &  0.379 & 0.281 & 0.184 & 0.694 & 0.643 & 0.820\\
    \rowcolor{gray!30} Sparse4Dv3 & EVA02-Large~\cite{fang2023eva02} & 8 & 0.668 & 0.346 & \textbf{0.279} & \textbf{0.142} & \textbf{0.719} & 0.677 & 0.761\\
    \bottomrule
  \end{tabular}}
  \caption{Experimental results of future frames and large backbone, where image size is set to $640 \times 1600$. "Future" represents the number of future frames used. TransFusion-L and -CL refer to pure lidar models and camera \& lidar multimodal models, respectively.}
  \label{tab:boost}
\end{table*}

\section{Conclusion and Outlook}
\label{Conclusion}
In this paper, we initially present methodologies to enhance the detection performance of Sparse4D. This enhancement primarily encompasses three aspects: Temporal Instance Denoising, Quality Estimation, and Decoupled Attention. Subsequently, we illustrate the process of expanding Sparse4D into an end-to-end tracking model.
Our experiments on the nuScenes show that these enhancements significantly improve performance, positioning Sparse4Dv3 at the forefront of the field. 

Based on the Sparse4D framework, there is considerable potential for further research:
(1) Our attempts at tracking are preliminary, and there is large room for improvement in tracking performance. (2) Expanding Sparse4D into a lidar-only or multi-modal model is a promising direction. (3) Building upon end-to-end tracking, further advancements can be made by introducing additional downstream tasks such as prediction and planning~\cite{uniad}. (4) Integrating additional perception tasks, such as online mapping~\cite{liao2022maptr} and 2D sign \& traffic light detection.

\newpage
{
\bibliographystyle{ieee_fullname}
\bibliography{egbib}
}


\end{document}